\PassOptionsToPackage{table}{xcolor}
\documentclass[11pt]{article}

\usepackage{acl}
\usepackage{times}
\usepackage{latexsym}
\usepackage[T1]{fontenc}
\usepackage[utf8]{inputenc}
\usepackage{microtype}
\usepackage{inconsolata}

\usepackage{graphicx}
\usepackage{booktabs}
\usepackage{tabularx}
\usepackage{amsmath, amssymb}
\usepackage{float}
\usepackage{multirow}
\usepackage{array}
\usepackage{subcaption}

\usepackage{tikz}
\usetikzlibrary{positioning, arrows.meta, calc, shapes.geometric, fit, backgrounds}
\usepackage{pgfplots}
\pgfplotsset{compat=1.18}

\title{Mitigating Bias in Automated Essay Scoring for ESL Learners via Contrastive Learning\thanks{Presented as a paper presentation at the 2026 AIME Conference in Pittsburgh, PA}}

\author{
  Kevin Fan \\
  Johns Creek High School \\\And
  Eric Yun \\
  Johns Creek High School \\
}

\begin{document}

\maketitle

\begin{abstract}
Automated Essay Scoring systems disproportionately penalize high-proficiency English as a Second Language (ESL) learners. We propose Contrastive Learning with Matched Essay Pairs (CL-MEP), a bi-directional alignment strategy. CL-MEP reduces this scoring bias by 39.9\% while improving overall accuracy, successfully disentangling valid syntactic complexity from surface-level grammatical errors.
\end{abstract}

\section{Introduction}
\label{sec:intro}

Automated Essay Scoring (AES) has transitioned from feature-engineering pipelines to transformer-based deep learning encoders, with models such as BERT \citep{devlin2019bert} and DeBERTa \citep{he2021deberta,he2023debertav3} now achieving human-competitive Quadratic Weighted Kappa (QWK) scores on standard benchmarks \citep{doewes2023qwk}. Despite strong aggregate performance, these models are prone to \emph{shortcut learning} \citep{geirhos2020shortcut}, where the model relies on easy-to-learn surface heuristics rather than complex semantic reasoning.

For ESL learners, this bias is especially harmful. Writers operating in a second language face elevated cognitive load, and their essays naturally exhibit L2 linguistic markers (e.g., prepositional misuse, non-standard clause structures, or atypically long sentences arising from complex L2 syntax) that are surface-level features distinct from underlying argumentation quality. Current AES models disproportionately attend to these markers as proxies for lower scores, effectively capping the achievable grade of ESL students regardless of actual writing quality. We term this phenomenon \emph{score compression bias}. Recent investigations indicate that AES systems encode demographic biases correlated with socioeconomic status \citep{yang2024unveiling}, a phenomenon that naturally extends to differing linguistic backgrounds.

While standard mitigation strategies such as adversarial debiasing exist \citep{zhang2018mitigating}, they often result in sharp accuracy trade-offs. Simple class re-weighting is also ineffective, as it forces the model to attend more heavily to the ESL class without correcting the underlying representational disparity \citep{amorim2018automated}. 

We suggest that \emph{representation-level} intervention is required. The encoder must learn to map semantic quality independently of linguistic origin. Contrastive learning has proven effective for debiasing pretrained text encoders \citep{cheng2021fairfil}, demonstrating that triplet-based objectives produce semantically rich embeddings \citep{gao2021simcse}. However, no prior work applies quality-controlled cross-group contrastive alignment specifically to AES fairness.

\paragraph{Contributions.} We make the following contributions:
\begin{enumerate}
  \item We conduct a systematic bias audit across two architectures (DeBERTa-v3, RoBERTa-base) and establish that score compression bias against high-proficiency ESL learners is architecture-inherent.
  \item We propose \textbf{CL-MEP}, a bi-directional matched-pair triplet construction strategy for quality-controlled cross-group contrastive alignment, providing ablations that demonstrate the necessity of bi-directionality.
  \item We show that CL-MEP achieves a 39.9\% reduction in high-proficiency bias \emph{while improving} QWK, incurring no accuracy tradeoff.
  \item We provide aggregate linguistic feature analysis over the full set of corrected essays, identifying sentence complexity as the primary mechanism of repair.
\end{enumerate}

\section{Related work}
\label{sec:related}

\paragraph{Fairness in Automated Essay Scoring.}
\citet{mayfield2019equity} establish foundational frameworks for equity in AES, distinguishing bias from variance and calling for demographic-aware evaluation. \citet{yang2024unveiling} provide a comprehensive recent audit, demonstrating that both BERT-based and feature-based AES models encode socioeconomic proxies. \citet{amorim2018automated} show that standard loss re-weighting is insufficient for correcting representational disparity in biased AES corpora, a finding our class re-weighting baseline replicates. Our work is the first to target the ESL/native speaker representational gap using contrastive alignment.

\paragraph{Contrastive Learning for NLP Debiasing.}
\citet{cheng2021fairfil} uses contrastive objectives to remove sensitive attribute information from pretrained encoder representations, demonstrating significant bias reduction on sentiment tasks. \citet{gao2021simcse} establishes that triplet margin loss produces superior sentence embedding uniformity. Our method differs from FairFil in that we do \emph{not} seek to erase group information; instead, we use quality-matched cross-group positives to teach the model that equal semantic quality should yield equal embedding proximity, regardless of L2 markers.

\paragraph{Adversarial Debiasing.}
\citet{zhang2018mitigating} introduce gradient-reversal adversarial training to remove protected-attribute information. While effective in classification settings, adversarial debiasing applied to regression-based AES degrades QWK substantially \citep{yang2024unveiling}, motivating our non-adversarial approach.

\section{Methodology}
\label{sec:method}

\subsection{Data preparation}
\label{sec:data}

We construct a unified fairness benchmark by merging two established corpora:
\begin{enumerate}
  \item \textbf{ASAP 2.0} \citep{crossley2025asap}: A corpus of argumentative essays (N$\approx$24,278) with metadata distinguishing native speakers (N$\approx$21,010) from ESL speakers (N$\approx$3,276).
  \item \textbf{ELLIPSE} \citep{crossley2023ellipse}: A corpus of $\approx$6,500 essays written exclusively by English Language Learners, with holistic proficiency scores.
\end{enumerate}

These datasets are complementary: ASAP~2.0 provides a native-majority corpus with an ESL minority for within-corpus comparison, while ELLIPSE provides a dense ESL dataset for triplet mining. Both utilize holistic scoring rubrics for argumentative writing. Scores are normalized to a floating-point range of [0, 1]. The combined dataset ($N = 30,778$) is split 80/20 stratified by group status and source corpus.

\subsection{Baseline model and hyperparameters}
\label{sec:baseline_model}

We fine-tune \texttt{microsoft/deberta-v3-base} using Low-Rank Adaptation (LoRA) \citep{hu2022lora} with rank $r=16$, $\alpha=32$, and dropout $0.1$ targeting the query, key, value, and dense projection layers. The model is trained with a regression head using Mean Squared Error (MSE) loss for 5 epochs, AdamW optimizer with learning rate $5\times10^{-5}$, cosine schedule with 10\% warmup, and effective batch size 16. All experiments are run with 3 random seeds $\{42, 123, 456\}$.

\subsection{Fairness triplet construction (CL-MEP)}
\label{sec:triplets}

\begin{figure}[t]
  \centering
  \resizebox{\columnwidth}{!}{%
  \begin{tikzpicture}[
      node distance=1.2cm and 2.0cm,
      box/.style={draw, rounded corners=4pt, minimum width=2.8cm,
                  minimum height=0.8cm, align=center, font=\small},
      anchorbox/.style={box, fill=blue!15},
      posbox/.style={box, fill=green!15},
      negbox/.style={box, fill=red!15},
      arrow/.style={-{Stealth[length=5pt]}, thick},
      label/.style={font=\footnotesize\itshape}
    ]
    \node[anchorbox] (A1) {Anchor (Native)\\score $S$};
    \node[posbox, right=of A1] (P1) {Positive (ESL)\\score $\in[S\!-\!0.02, S\!+\!0.02]$};
    \node[negbox, below=0.7cm of P1] (N1) {Negative (any)\\score $\notin[S\!\pm\!0.20]$};

    \node[anchorbox, below=1.8cm of A1] (A2) {Anchor (ESL)\\score $S$};
    \node[posbox, right=of A2] (P2) {Positive (Native)\\score $\in[S\!-\!0.02, S\!+\!0.02]$};
    \node[negbox, below=0.7cm of P2] (N2) {Negative (any)\\score $\notin[S\!\pm\!0.20]$};

    \node[draw=gray, dashed, rounded corners, fit=(A1)(P1)(N1), label={[font=\scriptsize]above:Direction 1}] {};
    \node[draw=gray, dashed, rounded corners, fit=(A2)(P2)(N2), label={[font=\scriptsize]above:Direction 2}] {};
  \end{tikzpicture}%
  }
  \caption{\textbf{Bi-directional CL-MEP triplet construction.} Each essay anchors two directions, doubling the training signal. Semantic quality, not origin, dictates representational proximity.}
  \label{fig:triplets}
\end{figure}

To implement contrastive learning, we algorithmically generate \emph{Fairness Triplets} $(A, P, N)$ using a bi-directional anchoring strategy (Figure~\ref{fig:triplets}). Triplets are generated after the train/test split to avoid data leakage.

\paragraph{Directionality.} Triplets are generated in two complementary directions:
\begin{enumerate}
  \item \textbf{Native $\to$ ESL:} For every native essay (anchor $A$) with score $S$, a score-matched ESL essay is the positive.
  \item \textbf{ESL $\to$ Native:} Symmetrically, every ESL essay is paired with a score-matched native positive, doubling the effective training signal.
\end{enumerate}

\paragraph{Threshold Design.}
\begin{itemize}
  \item \textbf{Positive} $P$: Essay from the \emph{opposite group} with human score $\in [S - 0.02,\; S + 0.02]$. The $\pm$0.02 bound ensures high-confidence semantic matching.
  \item \textbf{Negative} $N$: Any essay with score $\notin [S \pm 0.20]$, representing at least one full letter-grade equivalent of quality difference.
\end{itemize}

This construction yields $\approx$34,000 bi-directional triplets in total.

\subsection{Contrastive training and head re-attachment}
\label{sec:contrastive_training}

We remove the regression head and fine-tune the DeBERTa-v3 backbone using \textbf{Triplet Margin Loss}:
\begin{equation}
  \mathcal{L}_{\text{triplet}} = \max\!\bigl(0,\; d(A,P) - d(A,N) + \alpha\bigr),
  \label{eq:triplet}
\end{equation}
where $d(\cdot,\cdot)$ is Euclidean distance in the $[CLS]$ embedding space and margin $\alpha=1.0$.

After contrastive alignment, we employ a \emph{partial backbone unfreeze} strategy. The top 2 DeBERTa encoder layers and all LoRA parameters remain trainable, while the remaining backbone weights are frozen. A lightweight linear regression head is trained on top using MSE loss for 8 epochs. Separate learning rates are applied: $\text{LR}_{\text{head}} = 5\times10^{-4}$ and $\text{LR}_{\text{backbone}} = 5\times10^{-6}$ (a $100\times$ differential to preserve alignment).

\section{Experiments}
\label{sec:experiments}

\subsection{Evaluation metrics and competing methods}
\label{sec:metrics}

We evaluate \textbf{Accuracy} using macro-averaged Quadratic Weighted Kappa (QWK) per-source. For \textbf{Fairness}, we evaluate:
\begin{enumerate}
  \item \textbf{High-Proficiency Gap} (primary): Mean residual difference between Native and ESL essays with human score $> 0.8$. This captures the score compression effect at the top of the ability distribution.
  \item \textbf{Mean Residual by Group}: Global prediction error capturing directional bias.
  \item \textbf{Calibration Gap}: Difference in Expected Calibration Error.
\end{enumerate}

We compare against: (1) \textbf{DeBERTa-v3 Baseline}; (2) \textbf{RoBERTa-base Baseline}; (3) \textbf{Class Re-Weighting} (inversely proportional, upweighting ESL $\approx 6.5\times$); (4) \textbf{Adversarial Debiasing} \citep{zhang2018mitigating} ($\lambda = 0.1$); (5) \textbf{FairFil} \citep{cheng2021fairfil} (contrastive debiasing with uniform in-batch negatives, lacking quality-matched positives); (6) \textbf{CL-MEP Unidirectional} (ablating bi-directionality); and (7) \textbf{CL-MEP (Ours)}.

\subsection{Main results}
\label{sec:results}

\begin{table}[t]
  \centering
  \resizebox{\columnwidth}{!}{%
  \begin{tabular}{lccc}
    \toprule
    \textbf{Method} & \textbf{QWK} ($\uparrow$) & \textbf{Hi-Prof Gap} ($\downarrow$) & \textbf{Gap Reduction} \\
    \midrule
    RoBERTa-base Baseline         & $0.701 \pm 0.006$ & $0.203 \pm 0.009$ & - \\
    \midrule
    DeBERTa-v3 Baseline$^\dagger$ & $0.728 \pm 0.005$ & $0.103 \pm 0.005$ & - \\
    + Class Re-Weighting$^\dagger$ & $0.732 \pm 0.006$ & $0.094 \pm 0.006$ & 8.7\% \\
    + Adversarial Debiasing$^\dagger$ & $0.701 \pm 0.008$ & $0.071 \pm 0.007$ & 31.1\% \\
    + FairFil (no match)$^\dagger$ & $0.739 \pm 0.005$ & $0.081 \pm 0.005$ & 21.4\% \\
    + CL-MEP Unidir.$^\dagger$ & $0.741 \pm 0.005$ & $0.081 \pm 0.006$ & 21.4\% \\
    \rowcolor{blue!8}
    + \textbf{CL-MEP (Ours)}$^\dagger$ & $\mathbf{0.756 \pm 0.004}$ & $\mathbf{0.062 \pm 0.003}$ & \textbf{39.9\%} \\
    \bottomrule
  \end{tabular}%
  }
  \caption{\textbf{Main results.} High-Prof Gap is the mean residual difference (Native $-$ ESL) for essays with human score $> 0.8$. All values are mean $\pm$ std over 3 seeds. $\dagger$ denotes methods run on DeBERTa-v3.}
  \label{tab:main}
\end{table}

Table~\ref{tab:main} summarizes results across all methods. 

\paragraph{Score Compression Bias is Architecture-Inherent.} The RoBERTa-base baseline exhibits a high-proficiency gap of 0.203, nearly twice that of DeBERTa-v3. This confirms the bias is not an artifact of DeBERTa's attention mechanism, but a systemic issue in transformer AES.

\paragraph{Standard Mitigations are Insufficient.} Class re-weighting reduces the gap by only 8.7\%. Adversarial debiasing achieves a larger gap reduction (31.1\%) but degrades QWK substantially. FairFil without quality-matched positives achieves 21.4\% reduction, highlighting the importance of our matched-pair design.

\paragraph{CL-MEP Simultaneously Improves Fairness and Accuracy.} Our full method reduces the high-proficiency gap by 39.9\% while \emph{increasing} QWK from 0.728 to 0.756. Disentangling surface ESL markers from semantic quality provides the regression head with cleaner, more generalizable representations. 

\paragraph{Bi-Directionality is Essential.} Unidirectional CL-MEP achieves only 21.4\% gap reduction. The ESL $\to$ Native direction provides the complementary signal needed to fully re-orient the space.

\subsection{Ablation study}
\label{sec:ablation}

\begin{table}[t]
  \centering
  \resizebox{\columnwidth}{!}{%
  \begin{tabular}{lccc}
    \toprule
    \textbf{Configuration} & \textbf{QWK} & \textbf{Hi-Prof Gap} & \textbf{Reduction} \\
    \midrule
    Full CL-MEP ($\alpha=1.0$) & $0.756 \pm 0.004$ & $0.062 \pm 0.003$ & 39.9\% \\
    \quad $-$ partial unfreeze & $0.743 \pm 0.005$ & $0.063 \pm 0.004$ & 38.8\% \\
    \quad $-$ bi-direction & $0.741 \pm 0.005$ & $0.081 \pm 0.006$ & 21.4\% \\
    \quad $\alpha=0.5$ & $0.749 \pm 0.005$ & $0.074 \pm 0.005$ & 28.2\% \\
    \quad $\alpha=2.0$ (aggressive) & $0.718 \pm 0.007$ & $0.064 \pm 0.004$ & 37.8\% \\
    \quad $\delta_P = 0.05$ (loose match) & $0.748 \pm 0.006$ & $0.079 \pm 0.006$ & 23.3\% \\
    \quad $\delta_P = 0.01$ (tight match) & $0.751 \pm 0.005$ & $0.068 \pm 0.004$ & 34.0\% \\
    \bottomrule
  \end{tabular}%
  }
  \caption{\textbf{Ablation study.} Effect of triplet margin $\alpha$, directionality, and backbone unfreeze strategy.}
  \label{tab:ablation}
\end{table}

Table~\ref{tab:ablation} reports ablations. The margin $\alpha=1.0$ lies on the Pareto frontier. Aggressive alignment ($\alpha=2.0$) does not meaningfully improve the gap (37.8\%) but collapses semantic nuances, degrading QWK. Regarding match radius $\delta_P$, tighter matching ($\delta_P=0.01$) yields slightly lower gap reduction due to fewer available pairs, while looser matching introduces quality noise. Allowing the top two encoder layers to adapt (partial unfreeze) provides a marginal QWK improvement without affecting fairness.

\subsection{Threshold sensitivity}
\label{sec:threshold_sensitivity}

\begin{table}[t]
  \centering
  \resizebox{\columnwidth}{!}{%
  \begin{tabular}{lcccc}
    \toprule
    \textbf{Score Threshold} & \multicolumn{2}{c}{\textbf{Baseline Gap}} & \multicolumn{2}{c}{\textbf{CL-MEP Gap}} \\
    \cmidrule(lr){2-3}\cmidrule(lr){4-5}
    & Gap & \# Essays & Gap & \# Essays \\
    \midrule
    $> 0.70$ & $0.089 \pm 0.004$ & 1,842 & $0.053 \pm 0.003$ & 1,842 \\
    $> 0.75$ & $0.096 \pm 0.005$ & 1,204 & $0.057 \pm 0.003$ & 1,204 \\
    $> 0.80$ & $0.103 \pm 0.005$ & 731  & $0.062 \pm 0.003$ & 731  \\
    $> 0.85$ & $0.111 \pm 0.006$ & 389  & $0.067 \pm 0.004$ & 389  \\
    \bottomrule
  \end{tabular}%
  }
  \caption{\textbf{High-proficiency threshold sensitivity.} The compression effect is stable across threshold choices.}
  \label{tab:threshold_score}
\end{table}

Table~\ref{tab:threshold_score} confirms that the score compression effect is statistically consistent across all tested high-proficiency thresholds (0.70--0.85), ruling out threshold-dependence as a confound.

\section{Analysis and Discussion}
\label{sec:analysis}

\subsection{Linguistic feature analysis of corrected essays}
\label{sec:linguistic}

To identify mechanisms of bias reduction, we conduct post-hoc linguistic analysis over $N=214$ ``Success Story'' essays: ESL essays underscored by the baseline ($\hat{y}_{\text{base}} < y_{\text{human}} - 0.05$) but repaired by CL-MEP ($> +0.05$ lift).

\begin{table}[t]
  \centering
  \resizebox{\columnwidth}{!}{%
  \begin{tabular}{lcccc}
    \toprule
    \textbf{Feature} & \textbf{Corrected} & \textbf{All ESL} & \textbf{All Native} & \textbf{$p$-value} \\
    \midrule
    Sentence Length & $21.8 \pm 3.4$ & $16.2 \pm 4.1$ & $17.9 \pm 3.8$ & $< 0.001$ \\
    Type-Token Ratio & $0.71 \pm 0.06$ & $0.67 \pm 0.07$ & $0.69 \pm 0.06$ & $0.003$ \\
    Flesch-Kincaid & $11.2 \pm 1.8$ & $9.4 \pm 2.1$  & $10.1 \pm 1.9$ & $< 0.001$ \\
    Grammar Error (/100) & $4.2 \pm 1.9$  & $5.8 \pm 2.3$  & $2.1 \pm 1.1$  & $< 0.001$ \\
    Subordinate Clause & $0.38 \pm 0.08$ & $0.29 \pm 0.09$ & $0.32 \pm 0.08$ & $< 0.001$ \\
    \bottomrule
  \end{tabular}%
  }
  \caption{\textbf{Aggregate linguistic features} ($N=214$ ``Success Stories''). Differences vs.\ all ESL test set are significant ($p < 0.01$).}
  \label{tab:linguistic}
\end{table}

Table~\ref{tab:linguistic} reveals a clear profile. Essays repaired by the contrastive model have markedly longer sentences (21.8 vs.\ 16.2 words), higher syntactic complexity (Flesch-Kincaid grade 11.2 vs.\ 9.4), and higher subordinate clause ratios than the average ESL essay. Crucially, their grammatical error rate (4.2) is higher than the native average (2.1). The baseline model conflated complex L2 clause structures with grammatical errors. The contrastive model successfully unlearned this shortcut, recognizing high syntactic complexity as a valid stylistic feature.

\subsection{Why does CL-MEP improve both?}
The simultaneous improvement in QWK and bias reduction occurs because the baseline's shortcut reliance on L2 surface markers degrades \emph{both} fairness and accuracy. It displaces semantic content from the representation space with surface noise. By forcing the encoder to align semantically equivalent essays, CL-MEP removes this noise, yielding cleaner representations globally.

\subsection{Limitations and Societal Impact}
While CL-MEP significantly reduces the relative gap, it adopts a more conservative absolute scoring policy (systematic underscoring shift of $\approx -0.03$ globally). Future work should apply post-hoc score recalibration (e.g., Platt scaling) to restore absolute accuracy. Additionally, evaluating CL-MEP against Large Language Models (LLaMA, GPT-4) is an important direction. 

A systematic 10\% score penalty against ESL learners can have profound consequences in high-stakes testing (e.g., college admissions). CL-MEP provides a concrete, deployable mitigation strategy that does not require changes to inference infrastructure, needing only a modified training procedure.

\section{Conclusion}
\label{sec:conclusion}

We present CL-MEP, a contrastive learning framework for mitigating algorithmic bias against ESL learners in Automated Essay Scoring systems. By organizing training data into quality-matched bi-directional triplets, the encoder learns to align representations by semantic quality rather than linguistic origin. CL-MEP achieves a 39.9\% reduction in the high-proficiency scoring gap while improving overall QWK, demonstrating the fairness-accuracy trade-off is not inherent to this problem.


\end{document}